\documentclass[conference]{IEEEtran}
\IEEEoverridecommandlockouts

\usepackage{cite}
\usepackage{amsmath,amssymb,amsfonts}
\usepackage{graphicx}
\usepackage{textcomp}
\usepackage{xcolor}
\usepackage{booktabs}
\usepackage{tabularx}
\usepackage{bm}
\usepackage{multirow}
\usepackage{caption}
\usepackage{float}
\usepackage{tikz}
\usepackage{lettrine}

\def\BibTeX{{\rm B\kern-.05em{\sc i\kern-.025em b}\kern-.08em
    T\kern-.1667em\lower.7ex\hbox{E}\kern-.125emX}}

\DeclareRobustCommand{\Net}{FunKANLite}

\begin{document}

\setcounter{topnumber}{1}
\setcounter{dbltopnumber}{2}
\renewcommand{\topfraction}{0.9}
\renewcommand{\dbltopfraction}{0.9}
\renewcommand{\textfraction}{0.07}
\renewcommand{\dblfloatpagefraction}{0.7}
\setlength{\textfloatsep}{0.45\baselineskip plus 0.1\baselineskip minus 0.1\baselineskip}
\setlength{\dbltextfloatsep}{0.45\baselineskip plus 0.1\baselineskip minus 0.1\baselineskip}
\setlength{\floatsep}{0.35\baselineskip plus 0.1\baselineskip minus 0.1\baselineskip}
\setlength{\dblfloatsep}{0.35\baselineskip plus 0.1\baselineskip minus 0.1\baselineskip}
\setlength{\intextsep}{0.35\baselineskip plus 0.1\baselineskip minus 0.1\baselineskip}
\setlength{\abovecaptionskip}{0.25\baselineskip}

\bstctlcite{IEEEexample:BSTcontrol}

\title{Hardware-Aware Functional Kolmogorov-Arnold Networks
for Efficient Medical Image Enhancement and Segmentation}

\author{
  \IEEEauthorblockN{Mohammad Sadegh Sirjani\textsuperscript{1}}
  \IEEEauthorblockA{\textsuperscript{1}Department of Computer Science\\
  University of Texas at San Antonio, San Antonio, TX, USA\\
  mohammadsadegh.sirjani@utsa.edu}
}

\maketitle

\begin{abstract}
Functional Kolmogorov-Arnold Networks (FunKAN) achieve state-of-the-art accuracy on MRI Gibbs artifact removal and anatomical segmentation, but their 11.6\,M parameters and 8.7\,GFLOPs are too large for edge medical devices. We present \Net{}, a two-stage, hardware-aware compression of FunKAN for point-of-care use. \Net-TR reduces the spatial prior and replaces the ResBlock offset predictor with a depthwise-separable block. It has $1.9\times$ fewer parameters than FunKAN and no loss in accuracy. We then distill \Net-TR into \Net-ST, which lowers the Hermite basis rank, factorizes the spatial prior into a low-rank form, and halves the filter widths. \Net-ST has $5.6\times$ fewer parameters and $3.7\times$ fewer GFLOPs than FunKAN. It stays within $1.4$\,pp IoU of FunKAN on BUSI, GlaS, and CVC-ClinicDB, and reaches 33.95\,dB PSNR on IXI. On an NVIDIA Jetson Orin Nano and a Raspberry Pi~5, \Net-ST reduces energy per inference by up to $68\%$ and raises throughput by $2.9\times$.

\end{abstract}


\section{Introduction}
\label{sec:introduction}

Deploying diagnostic AI at the point of care requires models that are both accurate and compact.
Two tasks are central to this goal: MRI Gibbs artifact removal, which reduces the $k$-space truncation ringing that blurs fine anatomical detail~\cite{veraart2016gibbs,kellner2016gibbs}, and medical image segmentation, which automatically outlines lesions and structures in oncology, gastroenterology, and radiology~\cite{ronneberger2015unet,oktay2018attention,zhou2018unetpp,valanarasu2022unext,isensee2021nnunet,chen2021transunet,ma2024umamba}.
State-of-the-art models perform well on both tasks, but remain far too large for the memory and power limits of the edge hardware used in portable and bedside devices~\cite{kadhim2022deep,borys2023explainable,howard2017mobilenets,sandler2018mobilenetv2,tan2019efficientnet,scalcon2024jetson,mousavi2026cogadapt,mousavi2026mambagaze,sirjani2025fogscheduling,sirjani2025sdnplacement,ghaffari2025qteiot}.

These high-performing models mostly rely on convolutional or transformer backbones whose components are designed mainly by experiment, with little theoretical basis~\cite{ronneberger2015unet,he2016resnet,dosovitskiy2021vit}.
Kolmogorov-Arnold Networks (KANs)~\cite{liu2024kan} offer a more theory-based alternative: learnable single-variable functions built on the Kolmogorov-Arnold representation theorem~\cite{kolmogorov1957representations}, where orthogonal-polynomial variants such as ChebyKAN~\cite{ss2024chebykan} and HermiteKAN~\cite{seydi2024hermite} improve stability; medical variants such as U-KAN~\cite{li2025ukan}, UKAGNet~\cite{drokin2024ukagnet}, and Kolmogorov-Arnold Transformers~\cite{yang2025kat} extend KAN backbones to vision.
A shared limitation of these models is that they treat the input as an unordered set of scalars, ignoring the 2D spatial structure of images.
FunKAN~\cite{penkin2025funkan} addresses this by extending the Kolmogorov-Arnold theorem to function spaces: it treats each 2D feature map as an element of a Hilbert space and builds its inner functions from a Fourier expansion over Hermite basis functions.
This spatially-aware backbone achieves the best reported accuracy on MRI Gibbs removal (IXI~\cite{zhao2020gibbs}) and segmentation (BUSI~\cite{al2020dataset}, GlaS~\cite{valanarasu2021medical}, CVC-ClinicDB~\cite{bernal2015wm}), but at 11.6\,M parameters and 8.7\,GFLOPs it remains too large for edge deployment.

\begin{figure}[!t]
  \centering
  \includegraphics[width=\columnwidth]{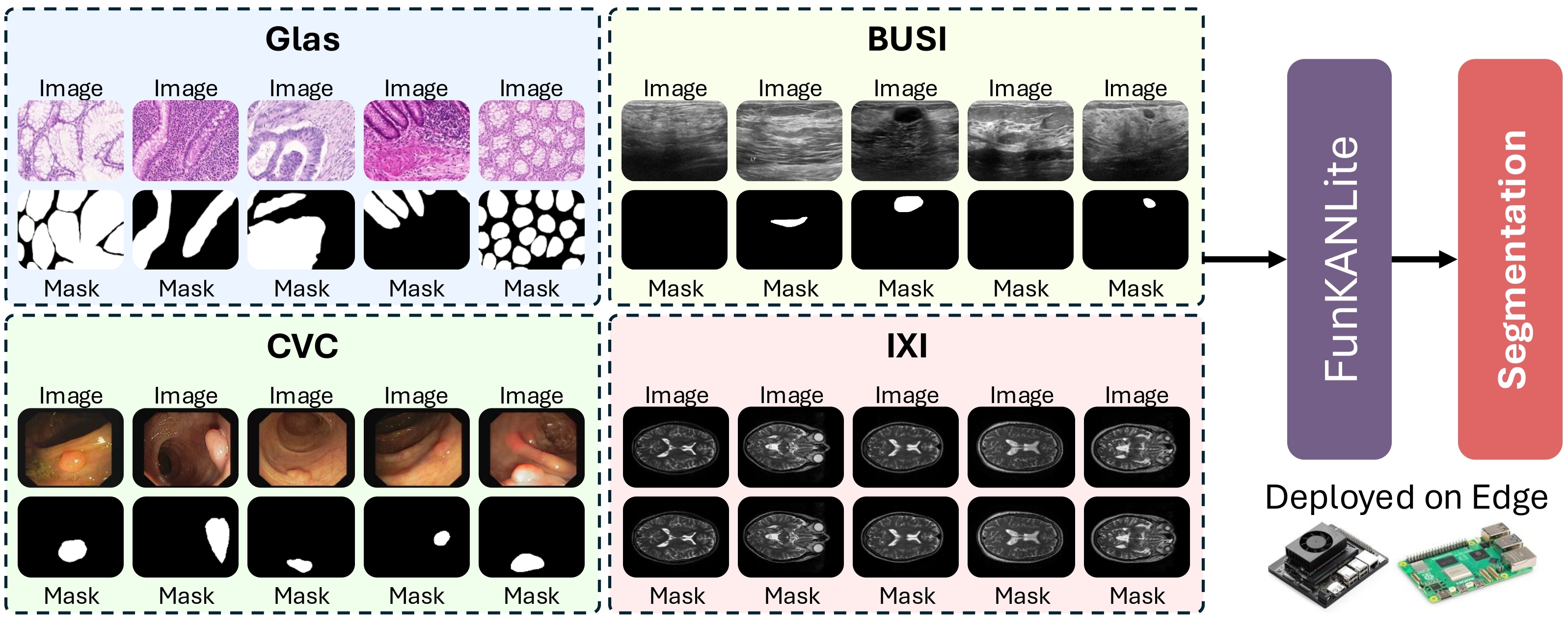}
  \caption{Overview of \Net{}. The network processes medical images and runs on an edge device for MRI Gibbs artifact suppression and anatomical segmentation.}
  \label{fig:overview}
\end{figure}

\begin{figure}[t]
  \centering
  \includegraphics[width=\columnwidth,height=0.34\textheight,keepaspectratio]{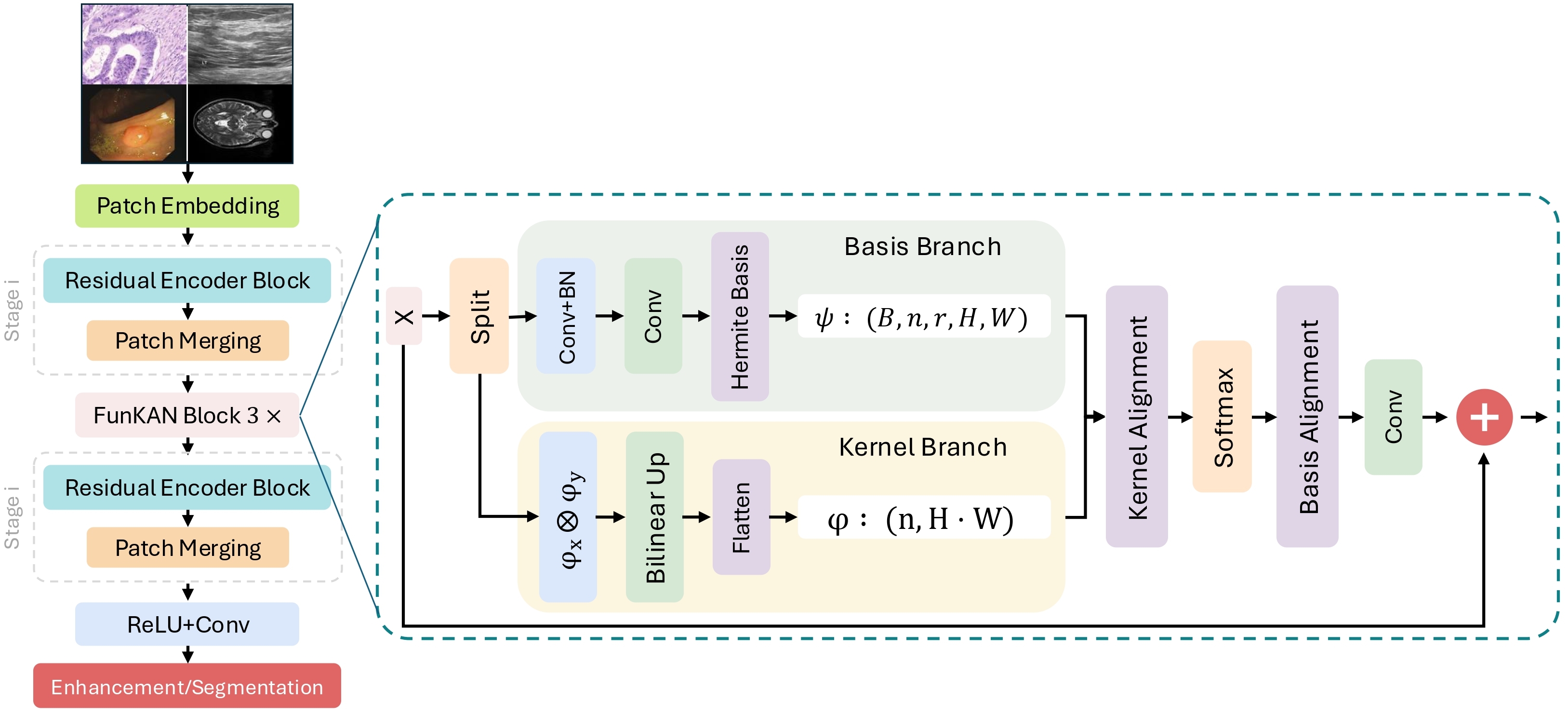}
  \caption{\Net{} architecture. The embedding, lifting, operator, projection, and restoration stages follow the FunKAN outer shell. Skip connections join the encoder and the decoder for segmentation.}
  \label{fig:architecture}
\end{figure}

\begin{figure}[t]
  \centering
  \includegraphics[width=\columnwidth,height=0.34\textheight,keepaspectratio]{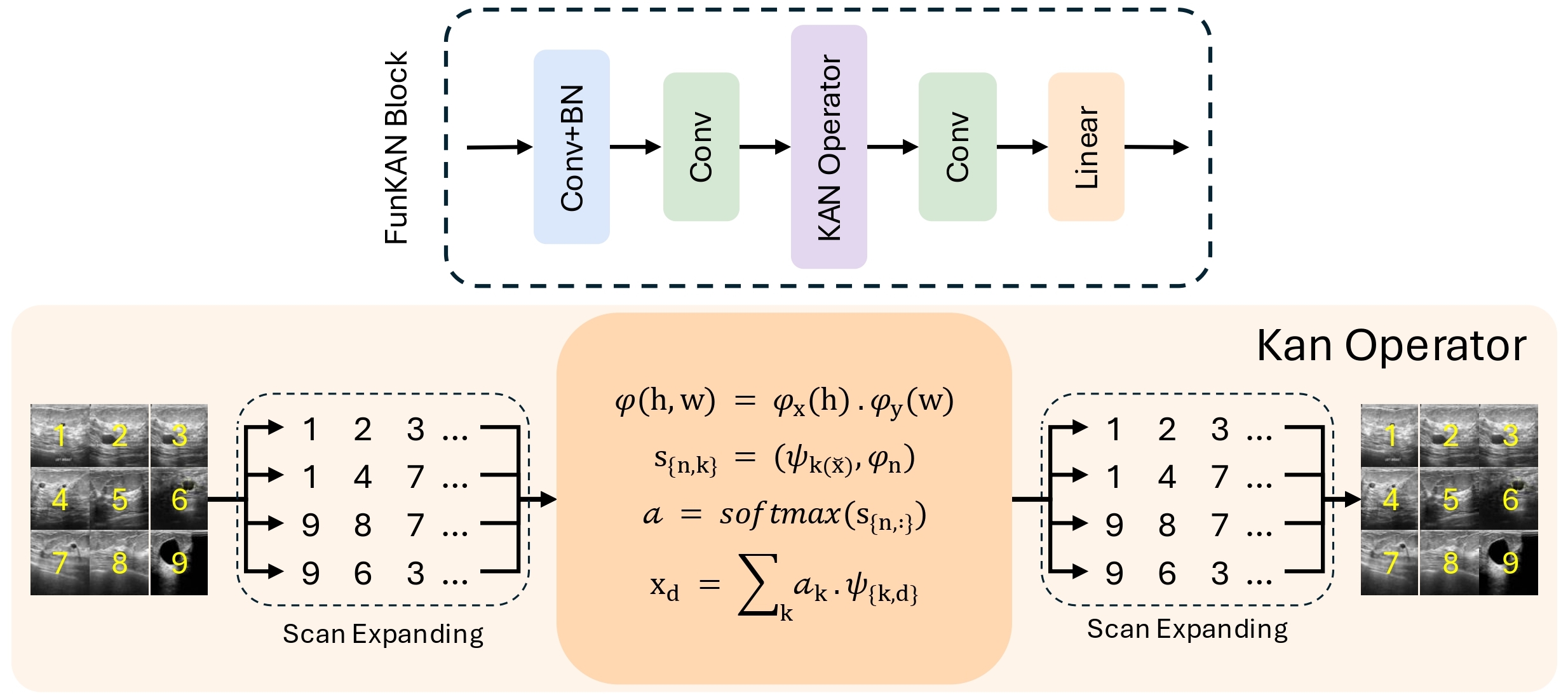}
  \caption{Operator block. A BN-ReLU-DW$_{3\times3}$-PW$_{1\times1}$ block replaces the ResBlock offset predictor of FunKAN (Eq.~\ref{eq:dwsep}). The spatial prior $\phi$ is stored at $110\!\times\!110$ and interpolated at runtime.}
  \label{fig:kanop}
\end{figure}

We present \Net{}, a hardware-aware compression of FunKAN for edge deployment, shown in Fig.~\ref{fig:overview}.
\Net-TR reduces the spatial prior from $145\!\times\!145$ to $110\!\times\!110$ and replaces the ResBlock offset predictor with a depthwise-separable block. It has $1.9\times$ fewer parameters than FunKAN and no drop in accuracy.
We then apply knowledge distillation~\cite{hinton2015distilling,romero2015fitnets,gou2021distillation} and train \Net-ST from \Net-TR. \Net-ST lowers the Hermite basis rank, factorizes the spatial prior into a low-rank form, and halves the filter widths. It has $5.6\times$ fewer parameters and $3.7\times$ fewer GFLOPs than FunKAN.
We measure both models on an NVIDIA Jetson Orin Nano and a Raspberry Pi~5.
Our contributions are:
\begin{itemize}
  \item We compress a functional KAN backbone for edge deployment. \Net{} keeps the spatially-aware representation of FunKAN.
  \item \Net-TR uses a compact spatial prior and a depthwise-separable offset predictor. It has $1.9\times$ fewer parameters than FunKAN and no accuracy loss. \Net-ST uses a lower Hermite rank, a low-rank prior, and halved widths. It has $5.6\times$ fewer parameters and $3.7\times$ fewer GFLOPs, and it stays within $1.4$\,pp IoU and $1.1$\,dB PSNR of FunKAN.
  \item We evaluate \Net-ST on an NVIDIA Jetson Orin Nano and a Raspberry Pi~5 across IXI, BUSI, GlaS, and CVC-ClinicDB. It cuts energy per inference by up to $68\%$ and raises throughput by $2.9\times$.
\end{itemize}

\section{Proposed Approach}
\label{sec:methodology}

\subsection{Architecture Design}

\begin{table}[t]
  \caption{Channel configuration at each encoder stage. E: embedding. S1--S4: encoder stages. The backbone uses the S4 width. MRI uses only E--S2.}
  \label{tab:kd}
  \centering
  \renewcommand{\arraystretch}{0.95}
  \begin{tabular*}{\columnwidth}{@{\extracolsep{\fill}} lccccc}
    \toprule
    Model & E & S1 & S2 & S3 & S4 \\
    \midrule
    FunKAN~\cite{penkin2025funkan} & 16 & 32 & 64  & 128 & 128 \\
    \Net-TR                        & 16 & 32 & 64  & 128 & 128 \\
    \Net-ST                        & 8  & 16 & 32  & 88  & 88  \\
    \bottomrule
  \end{tabular*}
\end{table}

\begin{figure}[t]
  \centering
  \includegraphics[width=0.78\columnwidth]{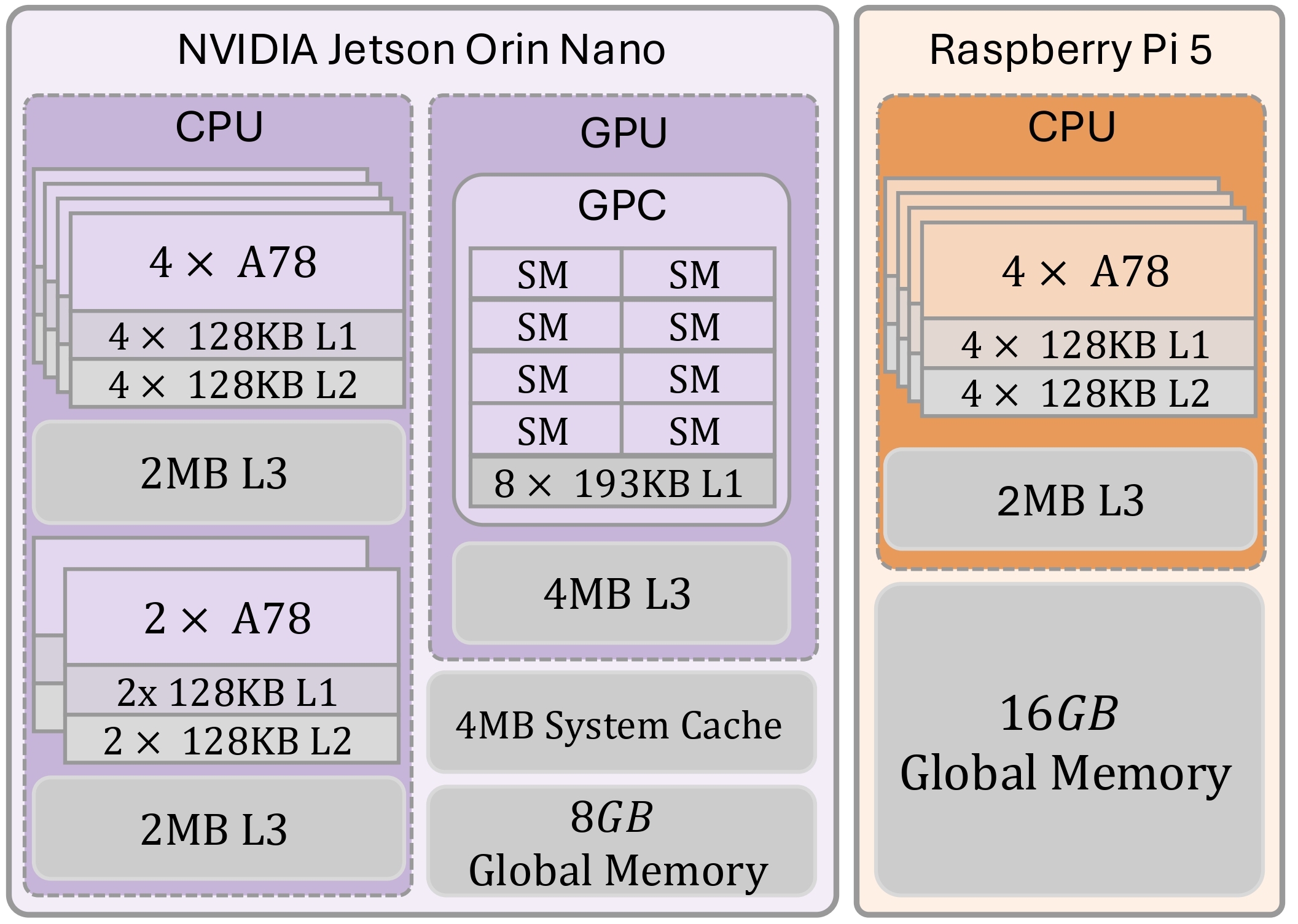}
  \caption{Hardware architecture of (a) NVIDIA Jetson Orin Nano and (b) Raspberry Pi~5. The diagram shows the L1, L2, and L3 caches and the on-chip memory used at inference.}
  \label{fig:hardware}
  \vspace{0.25em}
  \includegraphics[width=0.78\columnwidth]{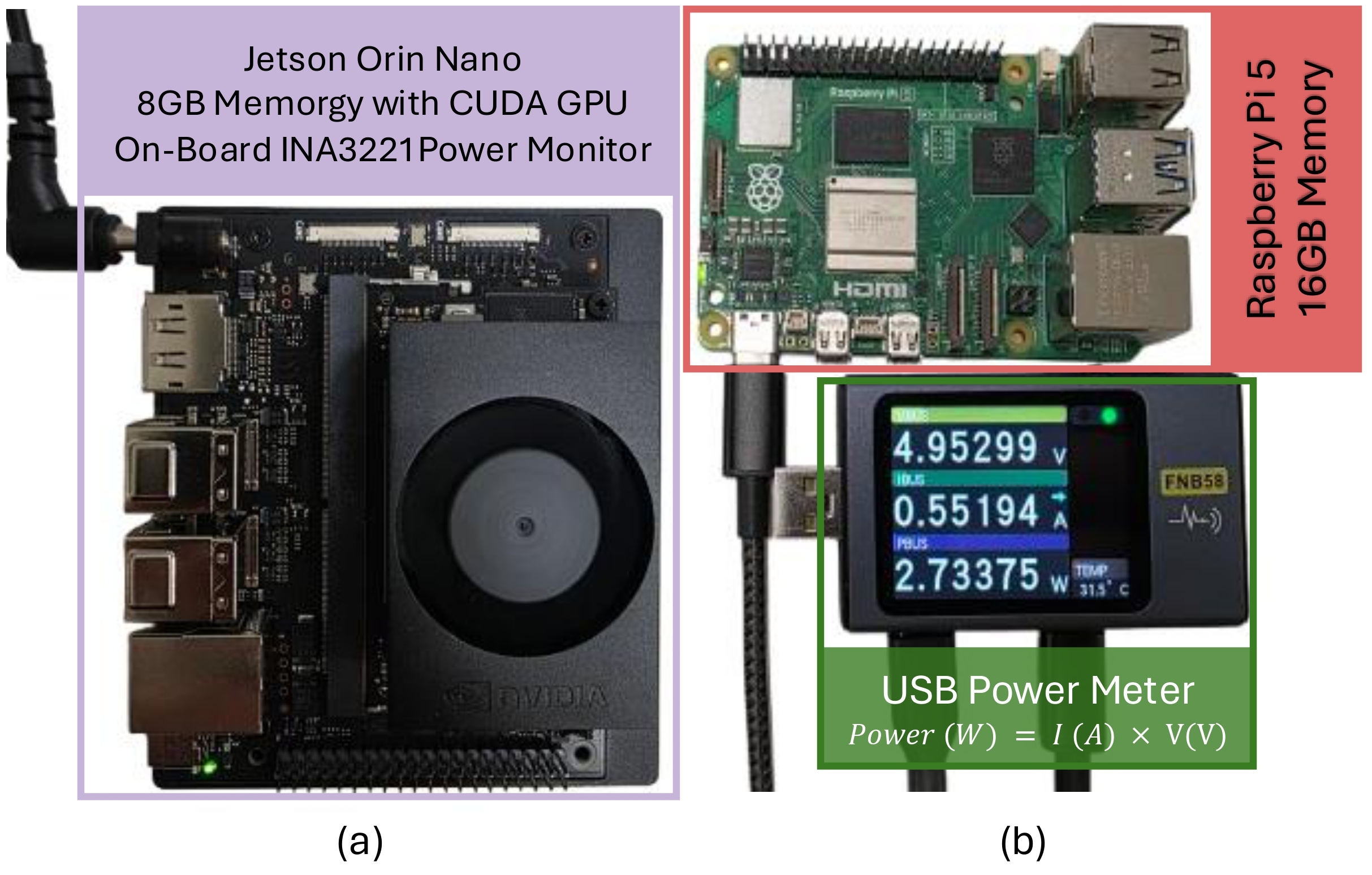}
  \caption{Power measurement setup. (a) NVIDIA Jetson Orin Nano with on-board INA3221 power monitor. (b) Raspberry Pi~5 with external FNB58 USB power meter.}
  \label{fig:power_setup}
\end{figure}

We realize the two-stage plan from Section~\ref{sec:introduction}. \Net-TR slims the FunKAN operator without losing accuracy. \Net-ST (Section~\ref{sec:st}) is distilled from \Net-TR and sized to the on-chip memory of the target hardware.
Fig.~\ref{fig:architecture} shows the architecture shared by \Net-TR and \Net-ST.
We adopt FunKAN~\cite{penkin2025funkan} as our baseline without changes, keeping its U-Net-style outer shell: a convolutional embedding, a stack of pre-activated residual encoder blocks~\cite{he2016resnet,duta2021improved} (lifting), three FunKAN-operator backbone blocks, a symmetric residual decoder with skip connections (projection), and a $1\!\times\!1$ restoration head.
For MRI enhancement, skip connections are removed and a $5\!\times\!5$ embedding is used. For segmentation, a $3\!\times\!3$ embedding and encoder-decoder skip connections are enabled.
All of our changes stay inside the FunKAN operator block. The outer network remains unchanged.

\subsubsection{FunKAN Operator}
As defined in~\cite{penkin2025funkan}, each backbone block evaluates the functional Kolmogorov-Arnold layer:
\begin{equation}
  \chi_{l+1,j} = \theta_{l,j}\!\left(\sum_{i=1}^{n} \sum_{k=1}^{r} \langle\varphi_{l,ji},\psi_k\rangle\,\psi_k(\chi_{l,i})\right),
  \label{eq:funkan}
\end{equation}
where $\{\psi_k\}_{k=1}^{r}$ is a bank of $r\!=\!6$ 2D Hermite basis functions, $\varphi_{l,ji}$ are the learned inner functions encoded in a spatial prior $\phi\!\in\!\mathbb{R}^{n\times 145\times 145}$, and $\theta_{l,j}$ is a $1\!\times\!1$ mixing convolution.
The inner products $\langle\varphi_{l,ji},\psi_k\rangle$ form an attention matrix that selects which Hermite modes are most relevant at each spatial location, which makes the representation easier to interpret.

The Hermite basis functions are evaluated on a grid that is deformed at runtime rather than a fixed uniform grid.
Given a base grid $\{q_x, q_y\}$, a residual offset predictor computes spatial displacements:
\begin{equation}
  \Delta q_l = \mathcal{W}_{l,0} \circ BN(\chi_l) + \mathcal{F}_l(\chi_l),
  \label{eq:offset}
\end{equation}
where $\mathcal{W}_{l,0}$ is a learned $3\!\times\!3$ convolutional layer and $\mathcal{F}_l$ is a two-layer pre-activated ResBlock(c $\to$ 2c, BN).
The deformed grid $q + \Delta q_l$ is then used to evaluate $\psi_k$, so the basis functions can adapt to local anatomy instead of remaining on a single global grid.
Backbone blocks are connected residually: $\chi_{l+1} = \chi_l + \text{FunKAN}(\chi_l)$.
This operator holds two parameter-heavy parts, the spatial prior $\phi$ and the offset predictor. \Net-TR compresses each of them.

\subsubsection{Compact Spatial Prior}
In FunKAN~\cite{penkin2025funkan}, the spatial prior $\phi$ is stored at $145\!\times\!145$ resolution and bilinearly interpolated to the current feature-map size $(H \times W)$ at each forward pass.
This resolution matches the MRI input and is larger than the compressed model needs.
\Net-TR stores the prior at $110\!\times\!110$. This cuts the $\phi$ parameter count per backbone block by $(145^2 - 110^2)/145^2 \approx 42\%$. Runtime interpolation keeps the spatial adaptivity.
Because $\phi$ is always interpolated before use, the attention computation in Eq.~\eqref{eq:funkan} is unchanged.

\subsubsection{Depthwise-Separable Offset Predictor}
The ResBlock offset predictor in FunKAN carries two full $3\!\times\!3$ convolutions (c $\to$ c $\to$ 2c), costing $\mathcal{O}(18c^2)$ parameters and proportional MACs per backbone block.
At the $128$-channel backbone stage used for segmentation, this accounts for a large share of the model size.
\Net-TR replaces the ResBlock with a depthwise-separable block~\cite{howard2017mobilenets}:
\begin{equation}
  \Delta q_l = \mathrm{PW}_{1\times1}\bigl(\mathrm{DW}_{3\times3}\bigl(\mathrm{ReLU}(BN(\chi_l))\bigr)\bigr),
  \label{eq:dwsep}
\end{equation}
where the depthwise $3\!\times\!3$ convolution operates independently per channel and the pointwise $1\!\times\!1$ convolution mixes channels to produce the $2c$ offset maps $(\Delta q_x, \Delta q_y)$.
This reduces the offset predictor to $\mathcal{O}(9c + 2c^2)$ parameters, about $20\times$ fewer than the ResBlock at large $c$, while preserving the same $3\!\times\!3$ receptive field that captures local deformation.

Fig.~\ref{fig:kanop} shows the resulting operator block.

\subsection{Memory-Aware \Net-ST}
\label{sec:st}

\Net-TR removes redundancy at no accuracy cost, but it remains too large for the edge. We therefore size \Net-ST to the memory limits of the target hardware.

\subsubsection{Memory Hierarchy Analysis}
Fig.~\ref{fig:hardware} illustrates the cache hierarchy of both target platforms.
The Jetson Orin Nano's Ampere GPU provides ${\sim}5.5$\,MB of on-chip memory ($8\!\times\!192$\,KB L1, $4$\,MB L2, $4$\,MB system-level cache); the Raspberry Pi~5, with no GPU, offers $4.5$\,MB on-chip via its Cortex-A76 L1/L2/L3 hierarchy and runs inference entirely on CPU.
A model whose working set fits in this budget avoids a main-memory transfer on every inference. The on-chip footprint is the compression target for \Net-ST.

\subsubsection{Architecture}
\Net-ST adds three reductions beyond \Net-TR. Each reduction targets a different source of model size.

\paragraph{Reduced Hermite basis rank}
The Hermite functions $\psi_k$ are ordered by increasing frequency: $\psi_0$ is a Gaussian, and higher-order modes capture progressively finer oscillatory patterns.
For smooth anatomical structures, most of the signal energy is concentrated in the low-order modes.
\Net-ST reduces the basis rank from $r\!=\!6$ to $r\!=\!4$ and drops $\psi_4$ and $\psi_5$. This reduces the attention matrix by $33\%$ and removes those Hermite evaluations.

\paragraph{Low-rank spatial prior}
The full spatial prior $\phi\!\in\!\mathbb{R}^{c\times H_\phi^2}$ stores $c \cdot H_\phi^2$ parameters per backbone block.
\Net-ST replaces it with a low-rank factorization:
\begin{equation}
  \phi \approx \mathbf{U}\mathbf{V}, \quad \mathbf{U}\!\in\!\mathbb{R}^{c\times\rho},\; \mathbf{V}\!\in\!\mathbb{R}^{\rho\times H_\phi^2},
  \label{eq:lowrank}
\end{equation}
with rank $\rho\!=\!40$ for segmentation and $\rho\!=\!12$ for MRI enhancement.
Both $\mathbf{U}$ and $\mathbf{V}$ are initialized with orthogonal matrices and learned end-to-end.
The product $\mathbf{U}\mathbf{V}$ is computed once per forward pass before the attention step. The extra cost is small, and $\phi$ storage falls by $(1 - \rho(c + H_\phi^2) / (c H_\phi^2))$. The reduction is large when $\rho \ll \min(c, H_\phi^2)$.

\paragraph{Halved filter widths}
Filter counts are reduced across all encoder stages, as detailed in Table~\ref{tab:kd}.
For segmentation, the backbone channel dimension drops from $128$ to $88$, and for MRI from $32$ to $16$.
Since the FunKAN operator complexity scales as $\mathcal{O}(c^2 \cdot r \cdot HW)$ through the attention einsum, this reduction has a quadratic effect on backbone compute.

\subsubsection{Knowledge Distillation Training}
We train \Net-ST from scratch and hold \Net-TR frozen.
The training objective combines three terms:
\begin{equation}
  \mathcal{L} = \alpha\,\mathcal{L}_{\text{task}}
              + \beta\,\mathcal{L}_{\text{feat}}
              + \gamma\,\mathcal{L}_{\text{out}},
  \label{eq:kd}
\end{equation}
with $\alpha\!=\!1.0$, $\beta\!=\!0.5$, and $\gamma\!=\!0.1$.

$\mathcal{L}_{\text{task}}$ is the primary supervised loss: MSE between predicted and clean images for MRI enhancement, and a weighted sum of binary cross-entropy and Dice loss for segmentation.

$\mathcal{L}_{\text{feat}}$ is a feature-level distillation term that aligns intermediate backbone representations:
\begin{equation}
  \mathcal{L}_{\text{feat}} = \frac{1}{|\mathcal{M}|}\sum_{l \in \mathcal{M}} \bigl\|f_l^{\mathrm{ST}} - f_l^{\mathrm{TR}}\bigr\|_2^2,
\end{equation}
where $\mathcal{M}$ is the set of backbone stages at which the two models share the same spatial size. Superscript $\mathrm{ST}$ denotes \Net-ST and $\mathrm{TR}$ denotes \Net-TR.
Stages with mismatched shapes (arising from the differing filter widths in Table~\ref{tab:kd}) are excluded automatically, avoiding the need for projection adapters.

$\mathcal{L}_{\text{out}}$ matches the final output of \Net-TR:
\begin{equation}
  \mathcal{L}_{\text{out}} = \bigl\|g^{\mathrm{ST}}(x) - g^{\mathrm{TR}}(x)\bigr\|_2^2,
\end{equation}
where $g^{\mathrm{ST}}$ and $g^{\mathrm{TR}}$ are the outputs of \Net-ST and \Net-TR.
This term complements the feature-level term by transferring the output of \Net-TR. The distillation temperature is $T\!=\!4.0$.

\begin{table*}[t]
  \caption{Per-dataset model complexity and hardware deployment on a Lambda server, a Jetson Orin Nano, and a Raspberry Pi~5.
    \Net-TR is the teacher. \Net-ST is the student.
    Thr.: throughput (img/s). Energy: energy per inference (mJ).
    The Lambda server GPU is not power-instrumented (Energy = N/A).}
  \label{tab:hardware}
  \centering
  \renewcommand{\arraystretch}{0.95}
  \scriptsize
  \begin{tabular*}{\linewidth}{@{\extracolsep{\fill}} ll rr rr rr rr}
    \toprule
    \multirow{2}{*}{Dataset} & \multirow{2}{*}{Model}
      & \multicolumn{2}{c}{Complexity}
      & \multicolumn{2}{c}{Lambda Server}
      & \multicolumn{2}{c}{Jetson Orin Nano}
      & \multicolumn{2}{c}{Raspberry Pi~5} \\
    \cmidrule(lr){3-4}\cmidrule(lr){5-6}\cmidrule(lr){7-8}\cmidrule(lr){9-10}
      & & Params\,(M) & GFLOPs & Thr. & Energy & Thr. & Energy & Thr. & Energy \\
    \midrule
    \multirow{3}{*}{BUSI} & FunKAN & 11.61 & 8.69 & 74.7  & N/A & 11.3 & 2893.6 & 5.8  & 1329.8 \\
                          & \Net-TR & 6.07  & 7.61 & 79.8  & N/A & 11.5 & 3264.6 & 6.2  & 1096.4 \\
                          & \Net-ST & 2.07  & 2.36 & 132.8 & N/A & 26.0 & 1425.5 & 16.3 & 444.9  \\
    \cmidrule(lr){1-10}
    \multirow{3}{*}{CVC}  & FunKAN & 11.61 & 8.69 & 72.3  & N/A & 11.9 & 2931.3 & 2.8  & 2062.5 \\
                          & \Net-TR & 6.07  & 7.61 & 85.0  & N/A & 11.5 & 3049.2 & 3.0  & 1787.3 \\
                          & \Net-ST & 2.07  & 2.36 & 140.4 & N/A & 25.8 & 1279.8 & 8.3  & 628.4  \\
    \cmidrule(lr){1-10}
    \multirow{3}{*}{GlaS} & FunKAN & 11.61 & 8.69 & 73.1  & N/A & 11.1 & 2990.7 & 2.8  & 1945.8 \\
                          & \Net-TR & 6.07  & 7.61 & 72.5  & N/A & 11.5 & 2837.2 & 3.1  & 1756.9 \\
                          & \Net-ST & 2.07  & 2.36 & 138.9 & N/A & 23.6 & 1353.5 & 8.3  & 633.2  \\
    \cmidrule(lr){1-10}
    \multirow{3}{*}{IXI}  & FunKAN & 2.17  & 6.22 & 58.2  & N/A & 10.8 & 6428.6 & 2.4  & 2238.8 \\
                          & \Net-TR & 1.19  & 0.76 & 87.4  & N/A & 14.3 & 2591.8 & 11.7 & 446.4  \\
                          & \Net-ST & 0.44  & 0.20 & 181.4 & N/A & 35.7 & 868.0  & 34.6 & 145.2  \\
    \bottomrule
  \end{tabular*}
\end{table*}

\begin{table}[t]
\caption{Segmentation accuracy, kept unchanged. Values use seeds $50$, $100$, and $150$, aggregated over the final $50$ epochs. They are not results of the current protocol.}
\label{tab:seg}
\centering
\scriptsize
\renewcommand{\arraystretch}{0.95}
\begin{tabular*}{\columnwidth}{@{\extracolsep{\fill}} llcc}
\toprule
Dataset & Model & IoU\,(\%) & F1\,(\%) \\
\midrule
BUSI & FunKAN~\cite{penkin2025funkan} & $68.06 \pm 0.85$ & $78.61 \pm 0.67$ \\
& \Net-TR & $68.94 \pm 0.12$ & $79.49 \pm 0.19$ \\
& \Net-ST & $67.34 \pm 0.46$ & $78.42 \pm 0.21$ \\
\midrule
CVC & FunKAN~\cite{penkin2025funkan} & $83.24 \pm 0.51$ & $89.72 \pm 0.46$ \\
& \Net-TR & $83.19 \pm 0.91$ & $89.49 \pm 0.71$ \\
& \Net-ST & $81.84 \pm 0.18$ & $88.80 \pm 0.10$ \\
\midrule
GlaS & FunKAN~\cite{penkin2025funkan} & $85.53 \pm 0.64$ & $91.93 \pm 0.41$ \\
& \Net-TR & $85.20 \pm 0.39$ & $91.77 \pm 0.27$ \\
& \Net-ST & $84.24 \pm 0.58$ & $91.17 \pm 0.35$ \\
\bottomrule
\end{tabular*}
\end{table}

\begin{table}[t]
\caption{Image-enhancement accuracy on IXI, kept unchanged. Same seeds and epoch window as Table~\ref{tab:seg}.}
\label{tab:enh}
\centering
\scriptsize
\renewcommand{\arraystretch}{0.95}
\begin{tabular*}{\columnwidth}{@{\extracolsep{\fill}} lcc}
\toprule
Model & PSNR\,(dB) & TV\,($\downarrow$) \\
\midrule
FunKAN~\cite{penkin2025funkan} & $35.01 \pm 0.01$ & $1156.8 \pm 1.4$ \\
\Net-TR & $34.80 \pm 0.05$ & $1145.8 \pm 1.8$ \\
\Net-ST & $33.95 \pm 0.06$ & $1123.7 \pm 4.5$ \\
\bottomrule
\end{tabular*}
\end{table}
\section{Experimental Results}
\label{sec:experiment}

\subsection{Experimental Setup}
\label{sec:setup}

\paragraph{Scope.}
The accuracy values in this section come from an earlier training protocol, described next.
We keep those values unchanged, and we did not regenerate them.
They are not results under the current protocol.

\paragraph{Implementation.}
The runs behind these tables used PyTorch Lightning and Adam with learning rate $10^{-4}$, batch size $8$, and up to $1000$ epochs.
Segmentation used Dice loss with weight $1.0$ plus binary cross-entropy with weight $0.1$.
Image enhancement used mean squared error.
\Net-TR is frozen during distillation.
Each reported accuracy is aggregated over the final $50$ epochs, for seeds $50$, $100$, and $150$.

\paragraph{Current protocol.}
The current segmentation recipe follows the U-KAN setup: $400$ epochs, Adam with weight decay $10^{-4}$, cosine annealing from $10^{-4}$ to $10^{-5}$, and learning rate $10^{-2}$ for the applicable KAN \texttt{layer} and \texttt{fc} parameters.
Its loss is Dice loss with smoothing $10^{-5}$ plus $0.5$ times binary cross-entropy.
IXI keeps the separate $1000$-epoch U-FunKAN learning-rate schedule.
Current segmentation runs use data-split seeds $2981$, $6142$, and $1187$, a fixed model RNG of $1029$, and the checkpoint with the best validation IoU.
That checkpoint is evaluated on the same held-out subset, so the number is a validation estimate.
IXI keeps its ordered test set and selects the checkpoint by validation loss.

\paragraph{Datasets.}
Segmentation is evaluated on BUSI~\cite{al2020dataset}, CVC-ClinicDB~\cite{bernal2015wm}, and GlaS~\cite{valanarasu2021medical} at $256\!\times\!256$ using IoU and F1.
Image enhancement is evaluated on IXI~\cite{zhao2020gibbs} at $145\!\times\!145$, with $k$-space truncation to $25\%$, using PSNR and TV.

\paragraph{Edge platforms.}
We evaluate on three platforms: a Lambda server, an NVIDIA Jetson Orin Nano, and a Raspberry Pi~5~\cite{scalcon2024jetson}, shown in Fig.~\ref{fig:power_setup}.
The Jetson runs PyTorch with CUDA. The Raspberry Pi~5 uses ONNX Runtime.
Power is sampled with the on-board INA3221 monitor on the Jetson and with an external FNB58 USB meter on the Raspberry Pi~5.
Both platforms execute $200$ runs after $10$ warm-up iterations.

\subsection{Results}
\label{sec:evaluation}

FunKAN is too costly to distill from directly, so \Net-TR is the source model.
Table~\ref{tab:seg} reports segmentation accuracy and Table~\ref{tab:enh} reports IXI enhancement. Table~\ref{tab:hardware} reports model size and measured deployment on all three platforms.
\Net-TR has $1.9\times$ fewer parameters than FunKAN. It uses a compact spatial prior and a depthwise-separable offset predictor, keeps segmentation accuracy, and reaches $34.80$\,dB PSNR on IXI.
Distilling \Net-TR into \Net-ST yields $5.6\times$ fewer parameters and $3.7\times$ fewer GFLOPs than FunKAN, at a cost of at most $1.4$\,pp IoU and $0.9$\,pp F1.
On a Jetson Orin Nano and a Raspberry Pi~5, \Net-ST fits in the on-chip cache. It provides up to $2.9\times$ higher throughput and $68\%$ lower energy per inference than FunKAN.

\section{Conclusion}
\label{sec:conclusion}

We presented \Net{}, a two-stage compression of FunKAN for edge medical imaging.
\Net-TR has $1.9\times$ fewer parameters than FunKAN and no accuracy loss, and distillation into \Net-ST extends this to $5.6\times$ fewer parameters and $3.7\times$ fewer GFLOPs, at a cost of at most $1.4$\,pp IoU.
On an NVIDIA Jetson Orin Nano and a Raspberry Pi~5, \Net-ST provides up to $2.9\times$ higher throughput and $68\%$ lower energy per inference.

\bibliographystyle{IEEEtran}
\bibliography{references}

\end{document}